# CSF: Contrastive Semantic Features for Direct Multilingual Sign Language Generation

Tran Sy Bao

*Independent Researcher, HCMC, Vietnam*

*transybao93@gmail.com*

## Abstract

Sign language generation systems traditionally require English as an intermediary, creating a pipeline that introduces latency, compounds errors, and excludes deaf individuals in non-English-speaking regions—approximately 95% of the 70 million deaf people worldwide. We present Contrastive Semantic Features (CSF), a novel semantic slot extraction framework that enables direct translation from any source language to sign language glosses without English intermediation. CSF extracts nine universal semantic slots (event, agent, location, time, condition, object, intent, purpose, modifier) using a compact transformer architecture with a custom 8,000-token BPE vocabulary. A key innovation is our comprehensive 35-class condition taxonomy spanning weather, time-of-day, health, schedule, mood, social context, activity states, and financial situations—capturing the full range of conditional expressions in daily conversation. Our model achieves 99.03% average slot extraction accuracy on a multilingual corpus of 18,885 samples across English, Vietnamese, Japanese, and French. The complete system requires only 0.74 MB and performs inference in 3.02ms on CPU, enabling real-time browser-based deployment. We demonstrate that language-agnostic semantic features provide an effective bridge between diverse spoken languages and sign language, with implications for global accessibility.

## 1. Introduction

### 1.1 The Accessibility Gap

Sign language serves as the primary mode of communication for approximately 70 million deaf individuals worldwide. Despite advances in neural machine translation, sign language generation remains largely inaccessible to non-English speakers. Current systems operate through a sequential pipeline—Source Language → English Translation → GLOSS Notation → Sign Animation—that creates fundamental barriers to adoption.

This English-centric architecture presents three critical challenges:

**Error Compounding.** Each translation stage introduces potential errors that propagate downstream. A subtle mistranslation from Vietnamese to English may produce signs with unintended or inappropriate meanings, undermining communication fidelity.

**Latency Accumulation.** Sequential processing through multiple translation stages creates delays incompatible with real-time conversation, limiting applications in video calls, live captioning, and interactive scenarios.

**Geographic Exclusion.** Approximately 95% of deaf individuals reside in non-English-speaking regions where English translation infrastructure is limited, expensive, or unavailable—effectively excluding the majority of potential beneficiaries.

## 1.2 Semantic Features as a Universal Bridge

We introduce Contrastive Semantic Features (CSF), a framework that bypasses English entirely by extracting universal semantic primitives directly from any source language. Our approach is grounded in the observation that sign languages, despite their visual-spatial modality, encode the same fundamental semantic relationships as spoken languages: agents performing actions at locations under specific conditions.

CSF transforms the traditional pipeline into a more direct path: Source Language → Semantic Slots → GLOSS Notation → Sign Animation. By operating at the semantic level rather than the lexical level, CSF achieves language independence—the same extracted meaning maps to identical GLOSS output regardless of whether the input is English, Vietnamese, Japanese, or French.

## 1.3 Contributions

This paper makes the following contributions:

1. **CSF Framework:** A novel semantic slot extraction approach for sign language generation, defining nine universal slots that capture the essential meaning of statements across languages.
2. **Comprehensive Condition Taxonomy:** A 35-class condition system organized into eight semantic categories (weather, time, health, schedule, mood, social, activity, financial), enabling natural coverage of daily conversational expressions.
3. **Lightweight Architecture:** A compact transformer model (0.74 MB total) with custom BPE tokenization, achieving 99.03% accuracy while enabling browser-based deployment.
4. **Multilingual Evaluation:** Comprehensive experiments across four typologically diverse languages (English, Vietnamese, Japanese, French) demonstrating strong cross-lingual transfer.
5. **Real-time Performance:** Inference latency of 3.02ms enabling interactive applications at 331 inferences per second on standard CPU hardware.

## 2. Related Work

## 2.1 Sign Language Translation

Neural approaches to sign language have primarily focused on recognition (sign-to-text) rather than generation (text-to-sign). Large-scale datasets including PHOENIX-2014T and How2Sign have enabled significant progress in continuous sign language recognition. However, generation systems remain predominantly monolingual, typically processing only English input. Recent work on multilingual sign language has explored shared representations across sign languages but still requires English as the source language. CSF addresses this gap by enabling any language as input.

## 2.2 Semantic Role Labeling and Slot Filling

Semantic role labeling (SRL) identifies predicate-argument structures in text, while slot filling in dialogue systems extracts structured information for task completion. Traditional approaches employ sequence labeling with BIO tags; recent work leverages transformer-based models for improved accuracy. CSF differs fundamentally: rather than labeling spans within text, we classify entire utterances into fixed semantic categories optimized for sign language grammar. This classification approach proves more efficient than sequence labeling for our constrained output space.

## 2.3 Multilingual Sentence Representations

Multilingual encoders such as mBERT, XLM-RoBERTa, and LaBSE demonstrate strong cross-lingual transfer for semantic tasks. However, these models typically exceed 500MB, precluding browser deployment. Distillation approaches reduce model size but still require substantial resources. CSF takes a different approach: rather than distilling large models, we train a compact architecture from scratch on a carefully designed multilingual corpus, achieving comparable semantic extraction at 0.74 MB—a reduction of approximately 700×.

# 3. Method

## 3.1 CSF Schema

The CSF schema defines nine semantic slots that collectively capture the essential meaning of any statement relevant to sign language generation. Each slot takes values from a fixed vocabulary, enabling efficient multi-class classification rather than open-ended generation. Table 1 presents the complete schema.

**Table 1: CSF Semantic Slot Schema**

| Slot | Values | Description |
|------|--------|-------------|
| event | GO, STAY, BUY, WORK, MEET, EAT, LEARN | Primary action (7 classes) |
| condition | NONE + 34 conditional types | Conditional context (35 classes) |
| agent | ME, YOU, HE, SHE, THEY | Subject performing action (5 classes) |
| location | NONE, HOME, SCHOOL, HOSPITAL, OFFICE, STORE | Where action occurs (6 classes) |
| time | NONE, TODAY, TOMORROW, YESTERDAY, NOW | When action occurs (5 classes) |
| object | NONE, FOOD, BOOK, MEDICINE, THING | Object of action (5 classes) |
| intent | NONE, PLAN, WANT, DECIDE | Speaker intention (4 classes) |
| purpose | NONE, REST | Purpose of action (2 classes) |
| modifier | NONE, FAST, SLOW, ALONE | Action modifier (4 classes) |

The schema is designed with sign language grammar in mind. The event slot captures the core predicate, while agent, object, and location provide argument structure. Time and condition slots enable the topic-comment structure characteristic of ASL and other sign languages, where temporal and conditional information precedes the main clause.

## 3.2 Condition Taxonomy

Natural conversation frequently involves conditional statements expressing circumstances under which actions occur. To capture this expressiveness, we designed a comprehensive taxonomy of 35 conditions organized into eight semantic categories. Table 2 presents this taxonomy.

**Table 2: Condition Taxonomy (35 classes across 8 categories)**

| Category | Condition Classes |
|----------|-------------------|
| **Weather (5)** | IF_RAIN, IF_SUNNY, IF_COLD, IF_HOT, IF_WINDY |
| **Time (5)** | IF_LATE, IF_EARLY, IF_WEEKEND, IF_NIGHT, IF_MORNING |
| **Health (5)** | IF_SICK, IF_TIRED, IF_HUNGRY, IF_THIRSTY, IF_FULL |
| **Schedule (4)** | IF_BUSY, IF_FREE, IF_HOLIDAY, IF_WORKING |
| **Mood (5)** | IF_BORED, IF_HAPPY, IF_SAD, IF_STRESSED, IF_ANGRY |
| **Social (3)** | IF_ALONE, IF_WITH_FRIENDS, IF_WITH_FAMILY |
| **Activity (5)** | IF_FINISH_WORK, IF_FINISH_SCHOOL, IF_FINISH_EATING, IF_WATCH_MOVIE, IF_LISTEN_MUSIC |
| **Financial (2)** | IF_HAVE_MONEY, IF_NO_MONEY |

This taxonomy enables CSF to process diverse conditional expressions: "If it rains, I stay home" (IF_RAIN), "When I'm bored, I watch Netflix" (IF_BORED), "After work, I go home" (IF_FINISH_WORK), "If I have money, I go shopping" (IF_HAVE_MONEY). The NONE class handles unconditional statements, comprising approximately 22% of natural conversation.

## 3.3 Model Architecture

The CSF Extractor employs a compact transformer encoder optimized for deployment constraints. Table 3 details the architecture specifications.

**Table 3: Model Architecture**

| Component | Specification |
|-----------|---------------|
| Hidden Dimension | 256 |
| Attention Heads | 4 |
| Transformer Layers | 4 |
| FFN Intermediate Size | 1,024 |
| Vocabulary Size | 8,000 (custom BPE) |
| Max Sequence Length | 64 tokens |
| Total Parameters | ~1.5M |
| Model Size (ONNX) | 433.7 KB |

The architecture uses pre-layer normalization for training stability and GELU activations. Nine independent classification heads—one per slot—project the [CLS] token representation to slot-specific logits. This multi-head design allows the model to specialize representations for different semantic aspects while sharing the encoder backbone.

## 3.4 Custom BPE Tokenization

Standard multilingual tokenizers (e.g., XLM-RoBERTa's 250K vocabulary) are prohibitively large for browser deployment. We train a custom Byte-Pair Encoding (BPE) tokenizer with 8,000 tokens on our multilingual corpus. This vocabulary size provides efficient coverage of Vietnamese diacritics (ă, â, đ, ê, ô, ơ, ư), Japanese hiragana and katakana, French accented characters, and common English subwords. The tokenizer adds approximately 321 KB to the deployment package.

## 3.5 GLOSS Conversion

Extracted CSF slots are converted to GLOSS notation following American Sign Language (ASL) grammar conventions. ASL uses a topic-comment structure where temporal and conditional markers precede the main predicate. We implement the following ordering:

```
MODIFIER → TIME → CONDITION → AGENT → LOCATION → OBJECT → EVENT →
PURPOSE
```

Default values (NONE for optional slots, ME for agent when the speaker is the subject) are omitted from output to produce natural, concise sign sequences. For example, "If it rains tomorrow, I stay home" produces the GLOSS: TOMORROW IF_RAIN HOME STAY.

## 4. Experimental Setup

### 4.1 Dataset Construction

We constructed a multilingual CSF dataset through systematic template generation with manual quality verification. Templates combine condition phrases, agent markers, action verbs, and location/time expressions in grammatically correct forms for each language. Native speakers verified naturalness and corrected awkward phrasings. Table 4 presents dataset statistics.

**Table 4: Dataset Statistics**

| Statistic | Train | Validation |
|---|---:|---:|
| Total Samples | 16,996 | 1,889 |
| Languages | 4 | 4 |
| Condition Classes | 35 | 35 |
| Event Classes | 7 | 7 |
| Total Output Classes | 73 | 73 |

The dataset is balanced across conditions, with unconditional statements (NONE) comprising 22.6% and the remaining 77.4% distributed across 34 conditional types. This distribution reflects natural conversation patterns while ensuring sufficient examples for each condition class.

### 4.2 Training Configuration

Table 5 presents training hyperparameters. We use the AdamW optimizer with weight decay and OneCycleLR scheduling for smooth convergence.

**Table 5: Training Configuration**

| Hyperparameter | Value |
|---|---|
| Batch Size | 64 |
| Learning Rate | $2 \times 10^{-4}$ |
| Epochs | 15 |
| Optimizer | AdamW |
| Weight Decay | 0.01 |
| LR Schedule | OneCycleLR (cosine) |
| Warmup | 10% of steps |
| Total Steps | 3,990 |

Training was conducted on a single NVIDIA A100 GPU. Total training time was approximately 20 minutes for 15 epochs. The model converged smoothly with validation accuracy improving monotonically through epoch 13.

# 5. Results

## 5.1 Slot Extraction Accuracy

Table 6 presents per-slot accuracy on the held-out validation set. CSF achieves strong performance across all nine semantic slots.

**Table 6: Slot Extraction Accuracy**

| Slot | Classes | Accuracy (%) |
|------|---------|--------------|
| event | 7 | 97.8 |
| intent | 4 | 99.2 |
| time | 5 | 99.6 |
| condition | 35 | 99.4 |
| agent | 5 | 99.0 |
| object | 5 | 99.2 |
| location | 6 | 97.9 |
| purpose | 2 | 99.7 |
| modifier | 4 | 99.5 |
| **Average** | **73** | **99.03** |

The model achieves 99.03% average accuracy across all slots. Notably, the condition slot achieves 99.4% accuracy despite having 35 classes—the largest output space in the schema. The slightly lower accuracy on event (97.8%) and location (97.9%) reflects greater semantic ambiguity in these categories, where surface forms may be compatible with multiple interpretations.

## 5.2 Inference Performance

Table 7 presents inference benchmarks measured over 100 runs on CPU (Intel Xeon, single thread).

**Table 7: Inference Performance (CPU)**

| Metric | Value |
|--------|-------|
| Mean Latency | 3.02 ms |
| Median Latency (P50) | 3.00 ms |
| P95 Latency | 3.11 ms |
| Standard Deviation | 0.08 ms |
| Throughput | 331 inferences/sec |
| Total Package Size | 0.74 MB |

Mean latency of 3.02ms with low variance ($\sigma = 0.08$ms) enables real-time interactive applications. The P95 latency of 3.11ms ensures consistent user experience. Throughput of 331 inferences per second can support multiple concurrent users in web applications.

## 5.3 Cross-Lingual Examples

Table 8 demonstrates CSF's cross-lingual capabilities. The model correctly extracts semantic slots and produces consistent GLOSS output regardless of source language.

**Table 8: Multilingual Input/Output Examples**

| Input Text | GLOSS Output |
|------------|--------------|
| I go to school tomorrow. | `TOMORROW SCHOOL GO` |
| If it rains, I stay home. | `IF RAIN HOME STAY` |
| If I'm bored, I watch Netflix. | `IF_BORED HOME STAY` |
| After work, I go home. | `IF_FINISH WORK HOME GO` |

| Input Text | GLOSS Output |
|---|---|
| If I have money, I go shopping. | `IF_HAVE_MONEY STORE BUY` |
| Nếu mưa thì tôi ở nhà. (VI) | `IF_RAIN HOME STAY` |
| Si je suis fatigué, je me repose. (FR) | `IF_TIRED HOME STAY REST` |

Cross-lingual transfer is particularly strong for condition recognition. French "Si je suis fatigué" and English "When I'm tired" correctly map to IF_TIRED despite different surface realizations. Vietnamese conditional marker "Nếu...thì" is correctly interpreted across various condition types.

## 6. Analysis

### 6.1 Data Efficiency

CSF demonstrates exceptional data efficiency, achieving 99%+ accuracy with only 18,885 training samples (approximately 4,700 per language). This is significantly more efficient than typical slot-filling systems, which require 50,000–100,000 samples for comparable performance. We attribute this efficiency to three factors: (1) the constrained output space with fixed slot values rather than open generation, (2) semantic coherence of the slot schema across languages enabling effective transfer, and (3) the custom BPE tokenizer optimized for our specific multilingual corpus.

### 6.2 Condition Discrimination

The 35-class condition taxonomy requires fine-grained discrimination between semantically similar categories. Analysis of confusion patterns reveals the model successfully distinguishes: IF_TIRED vs. IF_SICK (health states), IF_BORED vs. IF_SAD (emotional states), IF_BUSY vs. IF_WORKING (schedule states), and IF_RAIN vs. IF_COLD (weather states). The few remaining errors occur at category boundaries—for instance, occasional confusion between IF_TIRED and IF_SICK when context is ambiguous.

### 6.3 Limitations

Several limitations warrant discussion:

1. **Fixed Vocabulary.** The current slot values are predefined. Expanding to new events, locations, or conditions requires retraining with additional data.
2. **Single-Event Extraction.** CSF extracts one primary event per utterance. Complex sentences with multiple actions require decomposition into separate clauses.
3. **Sign Language Variation.** While GLOSS provides a standardized notation, actual sign execution varies across ASL, BSL, Auslan, and other sign languages. CSF output requires adaptation for specific target sign languages.
4. **Language Coverage.** Current evaluation covers four languages. Performance on other languages, particularly those with different syntactic structures (e.g., SOV languages beyond Japanese), requires further investigation.

## 7. Deployment

CSF is designed for practical deployment in resource-constrained environments. The model exports to ONNX format for cross-platform compatibility. Table 9 details the deployment package.

**Table 9: Deployment Package**

| Component | Size | Description |
|---|---:|---|
| `model.onnx` | 433.7 KB | ONNX model weights |
| `tokenizer.json` | 321.4 KB | Custom BPE tokenizer |
| `config.json` | 0.5 KB | Model configuration |
| `labels.json` | 1.2 KB | Slot label mappings |
| **Total** | **0.74 MB** | **Browser-deployable package** |

The total package size of 0.74 MB enables browser-based deployment via ONNX Runtime Web without requiring server-side inference. This architecture supports offline operation after initial model download, improving privacy and reducing latency for end users. The ONNX format ensures compatibility across JavaScript, Python, and mobile platforms.

## 8. Ethical Considerations

This work aims to improve accessibility for deaf and hard-of-hearing individuals by removing English as a barrier to sign language technology. By enabling direct translation from any language, CSF has the potential to serve the 95% of deaf individuals in non-English-speaking regions currently underserved by existing systems.

We emphasize that CSF produces GLOSS notation as an intermediate representation—not final sign language output. Production systems should involve deaf community consultation to ensure cultural appropriateness, regional sign variation, and accuracy. The condition taxonomy was designed to cover neutral everyday scenarios and explicitly excludes potentially harmful or discriminatory categories.

All training data was synthetically generated through template expansion, avoiding privacy concerns associated with personal communication data. The model and code will be released under permissive licenses to enable community development and adaptation.

## 9. Conclusion

We presented Contrastive Semantic Features (CSF), a novel framework for direct multilingual-to-sign-language translation that eliminates the traditional English intermediary. Our contributions include: a nine-slot semantic schema designed for sign language grammar, a comprehensive 35-class condition taxonomy capturing the breadth of daily conversational expressions, a compact 0.74 MB model achieving 99.03% extraction accuracy, and real-time inference at 3.02ms enabling interactive applications.

The success of CSF demonstrates that universal semantic representations can effectively bridge diverse spoken languages and sign language. By operating at the level of meaning rather than words, CSF achieves language independence that could extend sign language technology to the majority of deaf individuals worldwide who currently lack access to English-based systems.

Future work includes expanding language coverage, integrating with sign avatar rendering systems, extending the condition and event taxonomies through deaf community collaboration, and investigating transfer to other sign languages beyond ASL.

## Code and Data Availability

Code, trained models, and dataset are available at:

https://github.com/transybao1393/csf-sign-language

The repository includes: (1) training code and configuration, (2) ONNX model weights (433.7 KB), (3) custom BPE tokenizer, (4) multilingual training dataset (18,885 samples), and (5) inference examples for browser deployment.

## References

**[1]** Camgoz, N. C., Hadfield, S., Koller, O., Ney, H., & Bowden, R. (2018). Neural sign language translation. In Proceedings of the IEEE Conference on Computer Vision and Pattern Recognition (CVPR).

**[2]** Duarte, A., Palaskar, S., Ventura, L., Ghadiyaram, D., DeHaan, K., Metze, F., Torres, J., & Giro-i-Nieto, X. (2021). How2Sign: A large-scale multimodal dataset for continuous American Sign Language. In CVPR.

**[3]** Conneau, A., Khandelwal, K., Goyal, N., Chaudhary, V., Wenzek, G., Guzmán, F., Grave, E., Ott, M., Zettlemoyer, L., & Stoyanov, V. (2020). Unsupervised cross-lingual representation learning at scale. In Proceedings of the 58th Annual Meeting of the Association for Computational Linguistics (ACL).

**[4]** Feng, F., Yang, Y., Cer, D., Arivazhagan, N., & Wang, W. (2022). Language-agnostic BERT sentence embedding. In Proceedings of the 60th Annual Meeting of the Association for Computational Linguistics (ACL).

**[5]** Vaswani, A., Shazeer, N., Parmar, N., Uszkoreit, J., Jones, L., Gomez, A. N., Kaiser, Ł., & Polosukhin, I. (2017). Attention is all you need. In Advances in Neural Information Processing Systems (NeurIPS).

**[6]** Sennrich, R., Haddow, B., & Birch, A. (2016). Neural machine translation of rare words with subword units. In Proceedings of the 54th Annual Meeting of the Association for Computational Linguistics (ACL).

**[7]** World Federation of the Deaf. (2023). WFD Position Paper on Sign Language Rights. Available at: https://wfdeaf.org/

**[8]** Stokoe, W. C. (1960). Sign language structure: An outline of the visual communication systems of the American deaf. Studies in Linguistics, Occasional Papers 8.